\documentclass[runningheads]{llncs}
\usepackage{graphicx}
\usepackage{amsmath,amssymb} 
\usepackage{color}
\usepackage[width=122mm,left=12mm,paperwidth=146mm,height=193mm,top=12mm,paperheight=217mm]{geometry}

\usepackage[pagebackref=true,breaklinks=true,letterpaper=true,colorlinks,bookmarks=false]{hyperref}

\usepackage{times}
\usepackage{helvet}
\usepackage{courier}
\usepackage{graphicx}
\usepackage{bm}
\usepackage{amsmath,amssymb} 
\usepackage{color}
\usepackage{bbm}
\usepackage{epstopdf}
\usepackage{caption}
\usepackage{subcaption}
\usepackage{enumitem}
\usepackage{calc}
\usepackage{multirow}
\usepackage{xspace}
\usepackage{booktabs}
\usepackage{mathrsfs}
\usepackage{array}
\usepackage[font=small,skip=4pt]{caption}

\newcommand{\figref}[1]{Fig\onedot~\ref{#1}}
\newcommand{\equref}[1]{Eq\onedot~\eqref{#1}}
\newcommand{\secref}[1]{Sec\onedot~\ref{#1}}

\newcommand{\ve}[1]{{\mathbf #1}} 
\newcommand{\hua}[1]{{\mathcal #1}}
\newcommand{\scr}[1]{{\mathcal #1}}

\newcommand{\thickhline}{%
    \noalign {\ifnum 0=`}\fi \hrule height 1pt
    \futurelet \reserved@a \@xhline
}

\DeclareRobustCommand\onedot{\futurelet\@let@token\@onedot}
\def\onedot{\ifx\@let@token.\else.\null\fi\xspace}

\def\ie{\emph{i.e.}}

\def\etc{\emph{etc}\onedot} 

\def\wrt{w.r.t\onedot} 

\def\etal{\emph{et al.}}

\begin{document}
\pagestyle{headings}
\mainmatter

\title{Every Pixel Counts: Unsupervised Geometry Learning with Holistic 3D Motion Understanding} 

\titlerunning{Every Pixel Counts: Unsupervised Geometry Learning with Holistic 3D Motion Under}
\authorrunning{Zhenheng Yang~~Peng Wang~~Yang Wang~~Wei Xu~~Ram Nevatia}

\author{
Zhenheng Yang$^{1}$~~Peng Wang$^{2}$~~Yang Wang$^{2}$~~Wei Xu$^{3}$~~Ram Nevatia$^{1}$\\
}
\institute{
$^{1}$University of Southern California~~$^{2}$Baidu Research~~\\
$^{3}$National Engineering Laboratory for Deep Learning Technology and Applications\\
}

\maketitle

\begin{abstract}
    Learning to estimate 3D geometry in a single image by watching unlabeled videos via deep convolutional network has made significant process recently. Current state-of-the-art (SOTA) methods, are based on the learning framework of rigid structure-from-motion, where only 3D camera ego motion is modeled for geometry estimation. 
    However, moving objects also exist in many videos, \textit{e.g.} moving cars in a street scene.
    In this paper, we tackle such motion by additionally incorporating per-pixel 3D object motion into the learning framework, which provides holistic 3D scene flow understanding and helps single image geometry estimation. 
    Specifically, given two consecutive frames from a video, we adopt a motion network to predict their relative 3D camera pose and a segmentation mask distinguishing moving objects and rigid background. An optical flow network is used to estimate dense 2D per-pixel correspondence. A single image depth network predicts depth maps for both images. 
    The four types of information,  \ie~2D flow, camera pose, segment mask and depth maps, are integrated into a differentiable holistic 3D motion parser (HMP), where per-pixel 3D motion for rigid background and moving objects are recovered. We design various losses \wrt the two types of 3D motions for training the depth and motion networks, yielding further error reduction for estimated geometry.
    Finally, in order to solve the 3D motion confusion from monocular videos, we combine stereo images into joint training.
    Experiments on KITTI 2015 dataset show that our estimated geometry, 3D motion and moving object masks, not only are constrained to be consistent, but also significantly outperforms other SOTA algorithms, demonstrating the benefits of our approach.

\end{abstract}

\vspace{-0.6\baselineskip}
\section{Introduction}
\vspace{-0.1\baselineskip}
\label{sec:intro}

Humans are highly competent in recovering 3D scene geometry, \ie per-pixel depths, at a very detailed level. We can also understand both 3D camera ego motion and object motion from visual perception. 
In practice, 3D perception from images is widely applicable to many real-world platforms such as autonomous driving, augmented reality and robotics. This paper aims at improving both 3D geometry estimation from single image and also dense object motion understanding in videos.

\begin{figure}
\vspace{-0.8\baselineskip}
\includegraphics[width=\textwidth]{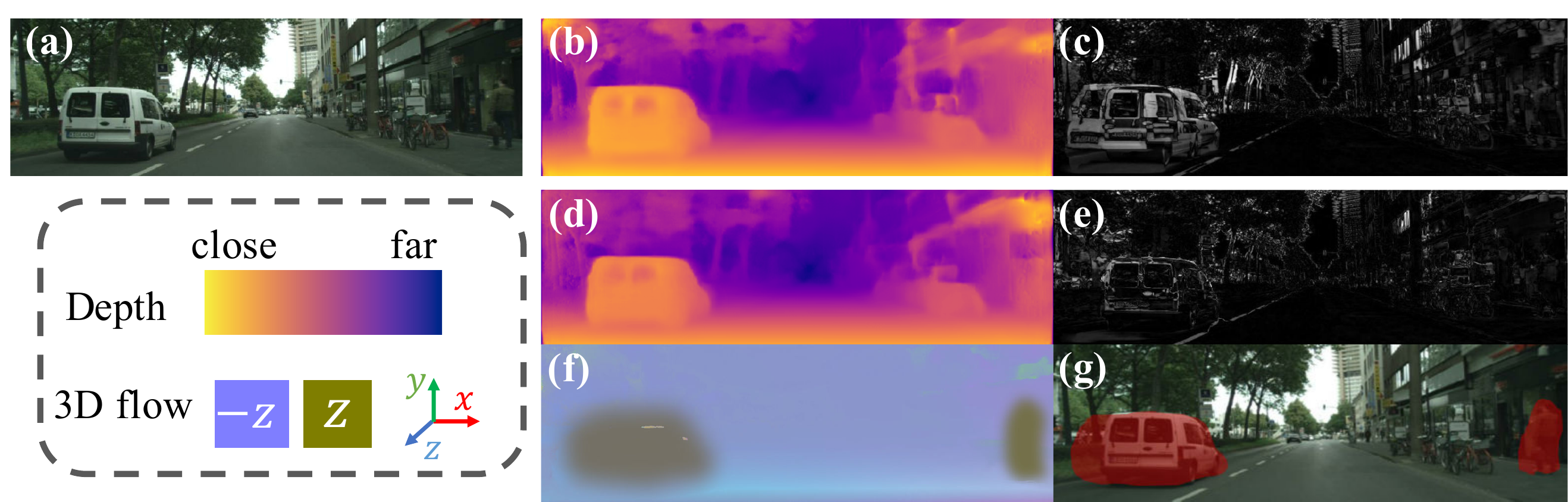}
\caption{With good depth estimation (b), there is still obvious reconstruction error around moving object (c). With joint training of 3D flow and depth, our framework generates depth result (d) and camera motion that causes less reconstruction error (e), and also consistent 3D scene flow (f) and moving object segmentation (g) results.}
\label{fig:example}
\vspace{-0.8\baselineskip}
\end{figure}

Recently, impressive progress~\cite{godard2016unsupervised,zhou2017unsupervised,yang2018aaai,yang2018cvpr} has been made to achieve 3D reconstruction from a single image by training a deep network taking only unlabeled videos or stereo images as input, yielding even better depth estimation results than those of supervised methods~\cite{eigen2014depth} in outdoor scenarios.
The core idea is to supervise depth estimation by view synthesis via rigid structure from motion (SfM)~\cite{wu2011visualsfm}. The frame of one view (source) is warped to another (target) based on the predicted depths of target view and relative 3D camera motions, and the photometric errors between the warped frame and target frame is used to supervise the learning. A similar idea also applies when stereo image pairs are available.
However, real world video may contain moving objects, which falls out of rigid scene assumption commonly used in these frameworks. As illustrated in \figref{fig:example}, with good camera motion and depth estimation, the synthesized image can still cause significant photometric error near the region of moving object, yielding unnecessary losses that cause unstable learning of the networks. 
Zhou \etal~\cite{zhou2017unsupervised} try to avoid such errors by inducing an explanability mask, where both pixels from moving objects and occluded regions from images are eliminated. 
Vijayanarasimhan \etal~\cite{Vijayanarasimhan17} separately tackle moving objects with a multi-rigid body model by outputting $k$ object masks and $k$ object pivots from the motion network.
However, such a system has limitations of maximum object number, and yields even worse geometry estimation results than those from Zhou \etal~\cite{zhou2017unsupervised} or other systems~\cite{yang2018cvpr} which do not explicitly model moving objects.

This paper aims for modeling the 3D motion for unsupervised/self-supervised geometry learning. Different from previous approaches, we model moving objects using dense 3D point offsets, \textit{a.k.a.} 3D scene flow, where the occlusion can be explicitly modeled. Thus, with camera motion in our model, every pixel inside the target image is explained and holistically understood in 3D.
We illustrate the whole model in \figref{fig:pipeline}. Specifically, given a target image and a source image, we first introduce an unsupervised optical flow network as an auxiliary part which produces two flow maps: from target to source and source to target images. Then, a motion network outputs the relative camera motion and a binary mask representing moving object regions, and a single view depth network outputs depths for both of the images. The four types of information (2D flow, camera pose, segment mask and depth maps) are fused with a holistic motion parser (HMP), where per-pixel 3D motion for rigid background and moving objects are recovered. 

Within the HMP, given depth of the target image, camera pose and moving object mask, a 3D motion flow is computed for rigid background. 
And given the optical flow, depths of the two images, an occlusion aware 3D motion flow of the full image is computed, where the occlusion mask is computed from optical flow following~\cite{wang2017occlusion}.
In principle, subtracting the two 3D flows within rigid regions, \ie without occlusion and outside moving object mask, the error should be zero. Inside moving object mask, the residual is object 3D motion, which should be spatially smooth. 
We use these two principles to guide additional losses formulation in our learning system, and all the operations inside the parser are differentiable. Thus, the system can be trained end-to-end, which helps the learning of both motion and depth.

For a monocular video, 3D depth and motion are entangled information, and could be confused with a projective camera model~\cite{torresani2008nonrigid}. For example, in the projective model, a very far object moving \wrt camera is equivalent to a close object keeping relatively still \wrt camera. The depth estimation confusion can be caused at regions of moving object.
We tackle this by also embedding the stereo image pair into the monocular learning framework when it is available. 
In our case, through holistic 3D understanding, we find the joint training yields much better results than solely training on stereo pairs or monocular videos individually.
Finally, as shown in \figref{fig:example}, our model successfully explains the optical flow to 3D motion by jointly estimating depths, understanding camera pose and separating moving objects within an unsupervised manner, where nearly all the photometric error is handled through the training process. Our learned geometry is more accurate and the learning process is more stable. 

We conduct extensive experiments over the public KITTI 2015~\cite{geiger2012we} dataset, and evaluate our results in multiple aspects including depth estimation, 3D scene flow estimation and moving object segmentation. As elaborated in \secref{sec:exp}, our approach significantly outperforms other SOTA methods on all tasks.

\vspace{-0.5\baselineskip}
\section{Related Work}
\vspace{-0.3\baselineskip}
\label{sec:related}
Estimating single view depth and predicting 3D motion from images have long been center problems for computer vision. Here we summarize the most related works in several aspects without enumerating them all due to space limitation.

\noindent\textbf{Structure from motion and single view geometry.}
Geometric based methods estimate 3D from a given video with feature matching or patch matching, such as PatchMatch Stereo~\cite{bleyer2011patchmatch}, SfM~\cite{wu2011visualsfm}, SLAM~\cite{mur2015orb,engel2014lsd} and DTAM~\cite{NewcombeLD11}, which could be effective and efficient in many cases. 
When there are dynamic motions inside a monocular video, usually there is scale-confusion for each non-rigid movement, thus regularization through low-rank~\cite{dai2014simple}, orthographic camera~\cite{taylor2010non}, rigidity~\cite{kumar2017monocular} or fixed number of moving objects~\cite{kumar2016multi} are necessary in order to obtain an unique solution. 
However, those methods assume 2D matching are reliable, which can fail at where there is low texture, or drastic change of visual perspective \etc. More importantly, those methods can not extend to single view reconstruction. 

Traditionally, specific rules are necessary for single view geometry, such as computing vanishing point~\cite{HoiemEH07}, following rules of BRDF~\cite{prados2006shape,kong2015intrinsic}, or extract the scene layout with major plane and box representations~\cite{DBLP:conf/iccv/SchwingFPU13,DBLP:conf/3dim/SrajerSPP14} \etc. These methods can only obtain sparse geometry representations, and some of them require certain assumptions (\textit{e.g.} Lambertian, Manhattan world).

\noindent\textbf{Supervised depth estimation with CNN.}
Deep neural networks (DCN) developed in recent years provide stronger feature representation. Dense geometry, i.e., pixel-wise depth and normal maps, can be readily estimated from a single image~\cite{wang2015designing,eigen2015predicting,laina2016deeper,li2017two} and trained in an end-to-end manner. The learned CNN model shows significant improvement compared to other methods based on hand-crafted features~\cite{karsch2014depth,ladicky2014pulling,zeisl2014discriminatively}. Others tried to improve the estimation further by appending a conditional random field (CRF)~\cite{DBLP:conf/cvpr/WangSLCPY15,Liu_2015_CVPR,li2015depth,peng2016depth}. However, all these methods require densely labeled ground truths, which are expensive to obtain in natural environments.

\noindent\textbf{Unsupervised single image depth estimation.}
Most recently, lots of CNN based methods are proposed to do single view geometry estimation with supervision from stereo images or videos, yielding impressive results. 
Some of them are relying on stereo image pairs~\cite{xie2016deep3d,GargBR16,godard2016unsupervised}, by warping one image to another given known stereo baseline. 
Some others are relying on monocular videos~\cite{zhou2017unsupervised,wang2017learning,li2017undeepvo,yang2018aaai,mahjourian2018unsupervised,yin2018geonet,yang2018cvpr} by incorporating 3D camera pose estimation from a motion network. However, as discussed in \secref{sec:intro}, most of these models only consider a rigid scene, where moving objects are omitted. 
Vijayanarasimhan \etal~\cite{Vijayanarasimhan17} model rigid moving objects with $k$ motion masks, while the estimated depths are negatively effected comparing to the one without object modeling~\cite{zhou2017unsupervised}. Yin \etal~\cite{yin2018geonet} model the non-rigid motion by introducing a 2D flow net, which helps the depth estimation. Different from those approaches, we propose to recover a dense 3D motion into the joint training of depth and motion networks, in which the two information are mutually beneficial, yielding better results for both depth and motion estimation. 



\noindent\textbf{3D Scene flow estimation.}
Estimating 3D scene flow~\cite{vedula2005three} is a task of finding per-pixel dense flow in 3D given a pair of images, which evaluates both the depth and optical flow quality. Existing algorithms estimate depth from stereo images~\cite{menze2015cvpr,behl2017bounding}, or the given image pairs~\cite{kumar2017monocular} with rigid constraint. And for estimation optical flow, they are trying to decompose the scene to piece-wise moving planes in order to finding correspondence with large displacement~\cite{vogel2013piecewise,lv2016continuous}.
Most recently, Behl \etal~\cite{behl2017bounding} adopt semantic object instance segmentation and supervised optical flow from DispNet~\cite{mayer2016large} to solve large displacement of objects, yielding the best results on KITTI dataset. 
Impressively, in our case, based on single image depth estimation and unsupervised learning pipeline for optical flow, we are able to achieve comparable results with the SOTA algorithms. This demonstrates the effectiveness of our approach.

\noindent\textbf{Segment moving objects.}
Finally, since our algorithm decomposes static background and moving objects, we are also related to segmentation of moving object from a given video. 
Current contemporary SOTA methods are dependent on supervision from human labels by adopting CNN image features~\cite{fragkiadaki2015learning,yoon2017pixel} or RNN temporal modeling~\cite{tokmakov2017learning}. 
For video segmentation without supervision, saliency estimation based on 2D optical flow is often used to discover and track the objects~\cite{wang2018saliency,faktor2014video,yang2017spatio}, and a long trajectory~\cite{brox2010object,kim2018face} of the moving objects needs to be considered. However, salient object is not necessary to be the moving object in our case. Moreover, we perform segmentation using only two consecutive images with awareness of 3D motion, which has not been considered in previous approaches. 

\section{Geometry Learning via Holistic 3D Motion Understanding}
\label{sec:approach}
As discussed in \secref{sec:intro}, a major drawback of previous approach~\cite{zhou2017unsupervised,yang2018cvpr} is ignorance of moving object. In the following, we will discuss the holistic understanding following the rule of geometry (\secref{subsec:scene_geometry}). Then, we elaborate how we combine stereo and monocular images with aware of 3D motion, and the losses used to train our depth networks.


\begin{figure*}[t]
\centering
\includegraphics[width=\textwidth]{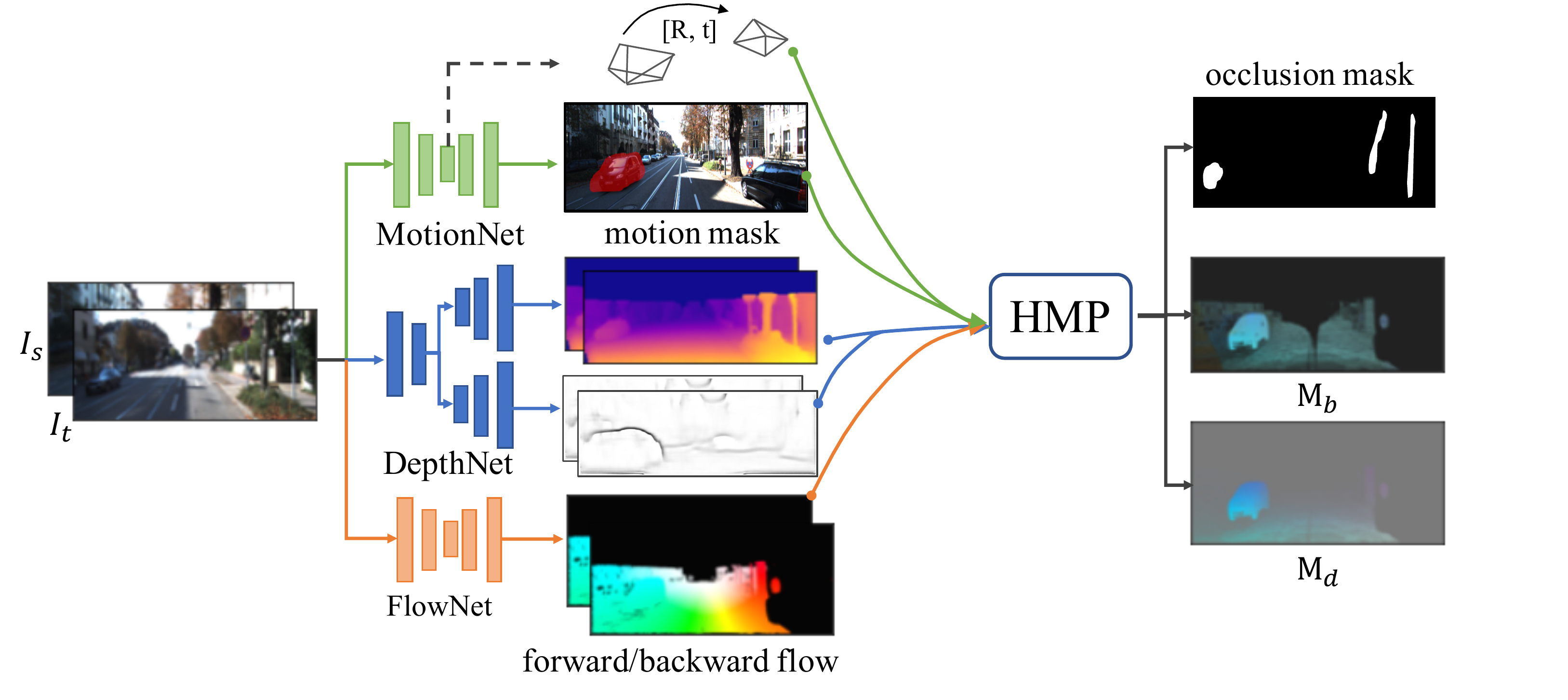}
\caption{Pipeline of our framework. Given a pair of consecutive frames, \ie target image $I_t$ and source image $I_{s}$, a FlowNet is used to predict optical flow $\ve{F}$ from $I_t$ to $I_{s}$. Notice here FlowNet is not the one in~\cite{IMKDB17}.
A MotionNet predicts their relative camera pose $\ve{T}_{t\rightarrow s}$ and a mask for moving objects $\ve{S}$. A single view DepthNet estimates their depths $\ve{D}_t$ and $\ve{D}_{s}$ independently. All the informations are put into our Holistic 3D Motion Parser (HMP), which produce an occlusion mask, 3D motion maps for rigid background $\ve{M}_s$ and dynamic objects $\ve{M}_d$. Finally, we apply corresponding loss over each of them.}
\label{fig:pipeline}
\vspace{-0.5\baselineskip}
\end{figure*}

\vspace{-0.5\baselineskip}
\subsection{Scene geometry with 3D motion understanding}
\label{subsec:scene_geometry}
 Given a target view image $I_t$ and a source view image $I_s$, suppose their corresponding depth maps are $\ve{D}_t, \ve{D}_s$, their relative camera  
 transformation is $T_{t\rightarrow s} = [\ve{R} | \ve{t}] \in \hua{S}\hua{E}(3)$ from $I_t$ to $I_s$, and 
 a per-pixel 3D motion map of dynamic moving objects $\ve{M}_d$ relative to the world.
 For a pixel $p_t$ in $I_t$, the corresponding pixel $p_s$ in $I_s$ can be found through perspective projection, \ie $p_s \sim \pi(p_t)$, 
\begin{align}
h(p_s) = \ve{V}(p_t)\frac{\ve{K}}{\ve{D}(p_s)}[\ve{T}_{t\rightarrow s}\ve{D}(p_t)\ve{K}^{-1}h(p_t) + \ve{M}_d(p_t)], 
\label{eqn:hmu}
\end{align}
where $\ve{D}(p_t)$ is the depth value of the target view at image coordinate $p_t$, and $\ve{K}$ is the intrinsic parameters of the camera, $h(p_t)$ is the homogeneous coordinate of $p_t$. $\ve{V}(p_t)$ is a visibility mask which is $1$ when $p_t$ is also visible in $I_s$, and $0$ if $p_t$ is occluded or flies out of image. In this way, every pixel in $I_t$ is explained geometrically in our model, yielding a holistic 3D understanding. Then given the corresponding $p_t$ and $p_s$, commonly, one may synthesize a target image $\hat{I}_t$ and compute the photometric loss $\|I_t(p_t) - \hat{I}_t(p_t)\|$ and use spatial transformer network~\cite{jaderberg2015spatial} for supervising the training of the networks~\cite{zhou2017unsupervised}.

Theoretically, given a dense matched optical flow from all available $p_t$ to $p_s$, when there is no non-rigid motion $\ve{M}$, \equref{eqn:hmu} is convex with respect to $\ve{T}$ and $\ve{D}$, and could be solved through SVD~\cite{tomasi1992shape} as commonly used in SfM methods~\cite{wu2011visualsfm}. This supports effective training of networks in previous works without motion modeling. 
In our case, $\ve{M}$ and $\ve{D}$ are two conjugate pieces of information, where there always exists a motion that can exactly compensate the error caused by depth. Considering matching $p_t$ and $p_s$ based on RGB could also be very noisy, this yields an ill-posed problem with trivial solutions.  Therefore, designing an effective matching strategies, and adopting strong regularizations are necessary to provide effective supervision for the networks, which we will elaborate later.

\noindent\textbf{Unsupervised learning of robust matching network.} As discussed in \secref{sec:related}, current unsupervised depth estimation methods~\cite{zhou2017unsupervised,wang2017learning,li2017undeepvo,mahjourian2018unsupervised,yang2018cvpr} are mostly based solely on photometric error, \ie $\|I_t(p_t) - \hat{I}_t(p_t)\|$, under Lambertian reflectance assumption and are not robust in natural scenes with lighting variations. More recently, supervision based on local structural errors, such as local image gradient~\cite{yang2018aaai}, and structural similarity (SSIM)~\cite{wang2004image,godard2016unsupervised,yin2018geonet} yields more robust matching and shows additional improvement on depth estimation. 

Structural matching has long been a center area for computer vision or optical flow based on SIFT~\cite{lowe2004distinctive} or HOG~\cite{lowe1999object} descriptors. Most recently, unsupervised learning of dense matching~\cite{wang2017occlusion} using deep CNN which integrates local and global context achieves impressive results 
according to the KITTI benchmark~\footnote{http://www.cvlibs.net/datasets/kitti/eval\_scene\_flow.php?benchmark=flow}. 
In our work, we adopt the unsupervised learning pipeline of occlusion-aware optical flow~\cite{wang2017occlusion} and a light-weighted network architecture, \ie PWC-Net~\cite{sun2017pwc}, to learn a robust matching using our training dataset. We found that although PWC-Net is almost 10$\times$ smaller than the network of FlowNet~\cite{IMKDB17} which was adopted by~\cite{wang2017occlusion}, it produce higher matching accuracy in our unsupervised setting. 


\noindent\textbf{Holistic 3D motion parser (HMP).} 
As described in \secref{sec:intro}, in order to apply the supervision, we need to distinguish between the motion from rigid background and 
dynamic moving objects. As illustrated in \figref{fig:pipeline}, we handle this through a HMP that takes multiple informations from the networks, and outputs the desired two motions. 

Formally, four information are input to HMP: depth of both images $\ve{D}_s$ and $\ve{D}_t$, the learned optical flow $\ve{F}_{t\rightarrow s}$, the relative camera pose $\ve{T}_{t\rightarrow s}$ and a moving object segment mask $\ve{S}_t$ inside $I_t$, where the motion of rigid background $\ve{M}_b$ and dynamic moving objects $\ve{M}_d$ are computed as,
\begin{align}
\ve{M}_b(p_t) &= \ve{V}(p_t)(1 - \ve{S}_t(p_t))[\ve{T}_{t\rightarrow s}\phi(p_t|\ve{D}_t) - \phi(p_t|\ve{D}_t)] \nonumber \\
\ve{M}_d(p_t) &= \ve{V}(p_t)\ve{S}_t(p_t)[\phi(p_t+\ve{F}_{t\rightarrow s}(p_t)|\ve{D}_{s}) - \phi(p_t|\ve{D}_t)]
\label{eqn:hmp}
\end{align}
where $\phi(p_t|\ve{D}_t) = \ve{D}_t(p_t)\ve{K}^{-1}h(p_t)$ is a back projection function from 2D to 3D space. $\ve{V}$ is the visibility mask as mentioned in \equref{eqn:hmu}, which could be computed by estimating an optical flow $\ve{F}_{s\rightarrow t}$ as presented in~\cite{wang2017occlusion}. We refer the reader to their original paper for further details due to the space limitation.

After HMP, the rigid and dynamic 3D motions are disentangled from the whole 3D motion, where we could apply various supervision accordingly based on our structural error and regularizations, which drives the learning of depth and motion networks.

\subsection{Training the networks.}\label{subsec:training}
In this section, we describe our loss design based on computed rigid and dynamic 3D motion from HMP. 
Specifically, as illustrated in \figref{fig:pipeline}, we adopt the network architecture from Yang \etal~\cite{yang2018cvpr}, which includes a shared encoder and two sibling decoders, estimating depth $\ve{D}$ and geometrical edge map $\ve{E}$ respectively, and a MotionNet estimating the relative camera poses. In this work, we also append a decoder with mirror connections in the same way with DepthNet to the MotionNet to output a binary segment mask $\ve{S}$ of the moving objects.

\noindent\textbf{Training losses.} 
Given background motion $\ve{M}_b(p_t)$ in \equref{eqn:hmp}, we can directly apply the structural matching loss by comparing it with our trained optical flow $\ve{F}_{t\rightarrow s}$ and the two estimated depth maps $\ve{D}_t, \ve{D}_s$ ($\hua{L}_{st}$ in \equref{loss:motion}). 
For moving objects $\ve{M}_d(p_t)$, we apply an edge-aware spatial smoothness loss for the motion map similar to that in~\cite{yang2018cvpr}. This is based on the intuition that motions belong to a single object should be smooth in real world ($\hua{L}_{ms}$ in \equref{loss:motion}).
Last, for $\ve{S}_t$ which segments the moving object, similar to the explainability mask in~\cite{zhou2017unsupervised}, we avoid trivial solutions of treating every pixel as part of moving objects by encouraging zeros predictions inside the mask (($\hua{L}_{vis}$ in \equref{loss:motion}).

In summary, the loss functions proposed in our work include,
\begin{align}
\hua{L}_{st} &= \sum\nolimits_{p_t}|\ve{M}_b(p_t) - \hat{\ve{M}}_b(p_t)|, \nonumber \\
\mbox{where,~~~} &\hat{\ve{M}}_b(p_t) = \ve{V}(p_t)(1 - \ve{S}_t(p_t))(\phi(p_t+\ve{F}_{t\rightarrow s}(p_t)|\ve{D}_{s}) - \phi(p_t|\ve{D}_t)), \nonumber \\
\scr{L}_{ms} &= \sum\nolimits_{p_t}(||\ve{M}_d(p_t)||^2+\sum\nolimits_{p_n \in \hua{N}_{p_t}}|\ve{M}(p_t) - \ve{M}(p_n)|\kappa(p_t, p_n|\ve{E}_t), \nonumber \\
\scr{L}_{vis} &= -\sum\nolimits_{p_t}\log(1 - \ve{S}_t(p_t))
\label{loss:motion}
\end{align}

where $\kappa(p_t, p_n|\ve{E}_t) = \exp\{-\alpha\max_{p\in\{p_t, p_n\}}(\ve{E}_t(p))\}$ is the affinity between two neighboring pixels, and $\hua{N}_{p_t}$ is a four neighbor set of pixel $p_t$, as defined in~\cite{yang2018cvpr}, which also helps to learn the EdgeNet.

In addition, in order to better regularize the predicted depths, we also add the depth normal consistency proposed in~\cite{yang2018aaai} for better regularization of depth prediction with normal information, and the losses corresponding to edge-aware depth and normal smoothness in the same way as~\cite{yang2018cvpr}, \ie $\hua{L}_D, \hua{L}_N$ and $\hua{L}_e$ respectively. We use $\scr{L}_{dne}$ to sum them up, and please refer to the original papers for further details. Here, different from~\cite{yang2018cvpr}, we apply such losses for both $\ve{D}_s$ and $\ve{D}_t$.

\noindent\textbf{Strong supervisions with bi-directional consistency.} Although we are able to supervise all the networks through the proposed losses in \equref{loss:motion}, we find that the training converges slower and harder when train from scratch compared to the original algorithm~\cite{yang2018cvpr}. The common solution to solve this is adding a strong supervision at the intermediate stages~\cite{simonyan2014very,lee2015deeply}. 
Therefore, we add a photometric loss without motion modeling for depth and camera motion prediction, and we apply the loss bi-directionally for both target image $I_t$ and source image $I_s$. Formally, our bi-directional view synthesis cost is written as,
\begin{align}
&\scr{L}_{bi-vs}\! =\! \sum\nolimits_{p_t}\!s(I_t(p_t), \hat{I}_t(p_t) | \ve{D}_t, \ve{T}_{t\rightarrow s}, I_s) + \sum\nolimits_{p_s}\!s(I_s(p_s), \hat{I_t}(p_s)| \ve{D}_s, \ve{T}_{s\rightarrow t}, I_t),  \nonumber \\
&\mbox{where,~~} s(I(p), \hat{I}(p) | \ve{D}, \ve{T}, I_s) = |I(p) - \hat{I}(p)| + \beta * \mbox{SSIM}(I(p), \hat{I}(p))
\label{eqn:bidirectional}
\end{align}
where the $\hat{I}_t(p)$ is the synthesized target image given $\ve{D}, \ve{T}, I_s$ in the same way with~\cite{zhou2017unsupervised}. $s(*, *)$ is a similarity function which includes photometric distance and SSIM~\cite{wang2004image}, and $\beta$ is a balancing parameter.

Finally, our loss functional for depth and motion supervision from a monocular video can be summarized as,
\begin{align}
\scr{L}_{mono}\! =\! \lambda_{st}\scr{L}_{st} + \lambda_{ms}\scr{L}_{ms} + \lambda_{vis}\scr{L}_{vis} + \sum\nolimits_l\{\lambda_{dne}\scr{L}_{dne}^l + \lambda_{vs}\scr{L}_{bi-vs}^l\}
\end{align}
where $l$ indicates the level of image resolution, and four scales are used in the same way with~\cite{zhou2017unsupervised}.

\begin{figure*}
\vspace{-0.5\baselineskip}
\centering
\includegraphics[width=\textwidth]{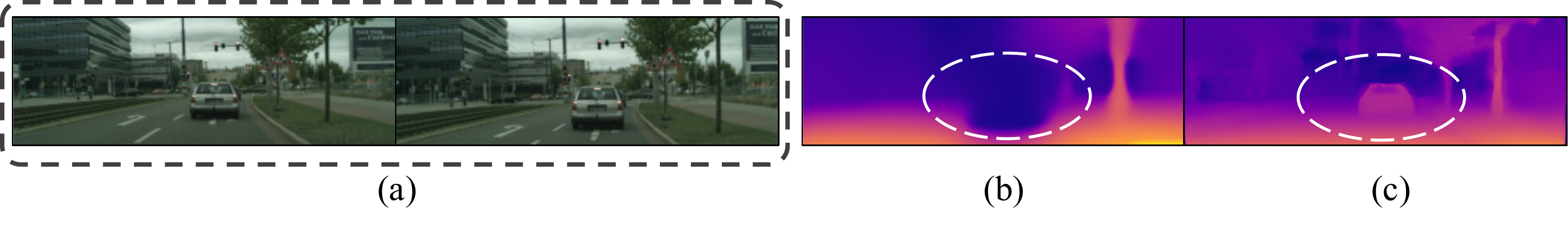}
\caption{Moving object in the scene (a) causes large depth value confusion for framework trained with monocular videos, as shown in (b). This issue can be resolved by incorporating stereo training samples into the framework (c). }
\label{fig:infinity}
\vspace{-0.5\baselineskip}
\end{figure*}

\noindent\textbf{Stereo to solve motion confusion.}
As discussed in our introduction (\secref{sec:intro}), reconstruction of moving objects in monocular video has projective confusion, which is illustrated in \figref{fig:infinity}. The depth map (b) is predicted with Yang \etal~\cite{yang2018cvpr}, where the car in the front is running at the same speed and the region is estimated to be very far. This is because when the depth is estimated large, the car will stay at the same place in the warped image, yielding small photometric error during training in the model. Obviously, adding motion or smoothness as before does not solve this issue. Therefore, we have added stereo images (which are captured at the same time) into learning the depth network to avoid such confusion. As shown in \figref{fig:infinity} (c), the framework trained with stereo pairs correctly figures out the depth of the moving object regions. 

Formally, when corresponding stereo image $I_{c}$ is additionally available for the target image $I_{t}$, we treat $I_{c}$ as another source image, similar to $I_{s}$, but with known camera pose $\ve{T}_{t\rightarrow c}$. In this case, since there is no motion factor, we adopt the same loss of $\hua{L}_{dne}$ and $\hua{L}_{bi-vs}$ taken $I_{c}, I_{t}$ as inputs for supervising the DepthNet. Formally, the total loss when having stereo images is, 
\begin{align}
\scr{L}_{mono-stereo}\! =\! \scr{L}_{mono} + \sum\nolimits_l\{\lambda_{dne}\scr{L}_{dne}^l(I_c) + \lambda_{vs}\scr{L}_{bi-vs}^l(I_c)\}.
\end{align}
where $\scr{L}_{dne}(I_c)$ and $\scr{L}_{bi-vs}(I_c)$ indicate the corresponding losses which are computed using stereo image $I_c$.

\vspace{-0.5\baselineskip}
\section{Experiments}
\vspace{-0.5\baselineskip}
\label{sec:exp}

In this section, we describe the datasets and evaluation metrics used in our experiments. And then present comprehensive evaluation of our framework on different tasks. 

\vspace{-0.3\baselineskip}
\subsection{Implementation details}
\vspace{-0.3\baselineskip}
Our framework consists of three networks: DepthNet, FlowNet and MotionNet. The DepthNet + MotionNet and FlowNet are first trained on KITTI 2015 dataset separately. Then DepthNet and MotionNet are further finetuned with additional losses from HMP as in Sec. \ref{sec:approach}.

\vspace{0.5\baselineskip}
\noindent\textbf{DepthNet architecture.}
A DispNet \cite{mayer2016large} like achitecture is adopted for DepthNet. Regular DispNet is based on an encoder-decoder design with skip connections and multi-scale side outputs. To train with stereo images, the output's channel for each scale is changed to 2, as in \cite{godard2016unsupervised}. As in \cite{yang2018cvpr}, the DepthNet has two sibling decoders which separately output depths and object edges. To avoid artifact grid output from decoder, the kernel size of decoder layers is set to be 4 and the input image is resized to be non-integer times of 64. All \textit{conv} layers are followed by ReLU activation except for the top output layer, where we apply a sigmoid function to constrain the depth prediction within a reasonable range. Batch normalization \cite{ioffe2015batch} is performed on all \textit{conv} layers. 
To increase the receptive field size while maintaining the number of parameters, dilated convolution with a dilation of 2 is implemented. 
During training, Adam optimizer \cite{kingma2014adam} is applied with $\beta_1=0.9$, $\beta_2=0.999$, learning rate of $2\times 10^{-3}$ and batch size of $4$. Other hyperparameters are set as in \cite{yang2018cvpr}.

\vspace{0.5\baselineskip}
\noindent\textbf{FlowNet architecture.}
A PWC-Net \cite{sun2017pwc} is adopted as FlowNet. PWC-Net is based on an encoder-decoder design with intermediate layers warping CNN features for reconstruction. 
The network is optimized with Adam optimizer \cite{kingma2014adam} with $\beta_1=0.9$, $\beta_2=0.999$, learning rate of $1\times 10^{-4}$ for 100,000 iterations and then $1\times 10^{-4}$ for 100,000 iterations. The batch size is set as 8 and other hyperparameters are set as in \cite{wang2017occlusion}.

\vspace{0.5\baselineskip}
\noindent\textbf{MotionNet architecture.}
The MotionNet implements the same U-net \cite{ronneberger2015u} architecture as the Pose CNN in \cite{zhou2017unsupervised}. The 6-dimensional camera motion is generated after 7 \textit{conv} layers and the motion mask is generated after symmetrical \textit{deconv} layers.

For end-to-end finetuning of DepthNet and MotionNet with HMP, the hyperparameters are set as: $\lambda_{st} = 0.5$, $\lambda_{ms} = 0.25$, $\lambda_{vis} = 0.8$, $\lambda_{dne} = 0.2$, $\lambda_{vs} = 1.0$. The trade-off weight between photometric loss and SSIM loss is set as $\beta=0.5$. All parameters are tuned on the validation set.

\vspace{-0.3\baselineskip}
\subsection{Datasets and metrics}
\vspace{-0.3\baselineskip}

Extensive experiments have been conducted on three different tasks: depth estimation, scene flow estimation and moving object segmentation. The results are evaluated on the KITTI 2015 dataset, using corresponding metrics.

\vspace{0.5\baselineskip}
\noindent\textbf{KITTI 2015.}
KITTI 2015 dataset provides videos in 200 street scenes captured by stereo RGB cameras, with sparse depth ground truths captured by Velodyne laser scanner. 2D flow and 3D scene flow ground truth is generated from the ICP registration of point cloud projection. The moving object mask is provided as a binary map to distinguish background and foreground in flow evaluation. During training, 156 stereo videos excluding test and validation scenes are used.The monocular training sequences are constructed with three consecutive frames in the left view, while stereo training pairs are constructed with left and right frame pairs, resulting in a total of 22,000 training samples. 

For depth evaluation, two test splits of KITTI 2015 are proposed: the official test set consisting of 200 images (KITTI split) and the test split proposed in \cite{eigen2014depth} consisting of 697 images (Eigen split). The official KITTI test split provides ground truth of better quality compared to Eigen split, where less than 5\% pixels in the input image has ground truth depth values. For better comparison with other methods, the depth evaluation is conducted on both splits. For scene flow and segmentation evaluation, as the flow ground truth is only provided for KITTI split, our evaluation is conducted on the 200 images in KITTI test split.

\vspace{0.5\baselineskip}
\noindent\textbf{Cityscapes.}
Cityscapes is a city-scene dataset captured by stereo cameras in 27 different cities. As depth ground truth is not available, Cityscapes is only used for training and the training samples are generated from 18 stereo videos in the training set, resulting in 34,652 samples.

\vspace{0.5\baselineskip}
\noindent\textbf{Metrics.} The existing metrics of depth, scene flow and segmentation have been used for evaluation, as in \cite{eigen2014depth}, \cite{menze2015cvpr} and \cite{long2015fully}. For depth and scene flow evaluation, we have used the code by \cite{godard2016unsupervised} and \cite{menze2015cvpr} respectively. 
For foreground segmentation evaluation, we implemented the evaluation metrics in \cite{long2015fully}. 
The definition of each metric used in our evaluation is specified in Tab. \ref{metrics}. In which, $x^*$ and $x'$ are ground truth and estimated results ($x \in \{d, sf\}$). $n_{ij}$ is the number of pixels of class $i$ segmented into class $j$. $t_i$ is the total number of pixels in class $i$. $n_{cl}$ is the total number of classes, which is equal to 2 in our case.


\begin{table}[!htbp]
\vspace{-1.8\baselineskip}
\centering
\fontsize{8}{10}\selectfont
\def\arraystretch{1.5}
\caption{From top row to bottom row: depth, scene flow and segmentation evaluation metrics.}
\setlength{\tabcolsep}{2pt}
\begin{tabular}{l|l}
\specialrule{.2em}{.1em}{.1em}
Abs Rel: $\!\frac{1}{|D|}\!\sum_{d'\in D}\!|d^*\!\!\!-\!\!d'|/d^*$       & Sq Rel: $\frac{1}{|D|}\!\sum_{d'\in D}\!||d^*\!\!\!-\!\!d'||^2\!/d^*$                \\
RMSE: $\!\sqrt{\!\frac{1}{|D|}\!\sum_{d'\!\in\! D}||d^*\!\!\!-\!\!d'||^2}$    & RMSE log: $\!\sqrt{\!\!\frac{1}{|D|}\!\!\sum_{d'\!\in\! D}\!\!||\!\log\! d^*\!\!\!-\!\!\log\! d'||^2\!}\!$ \\ \hline
D1, D2:  $\!\frac{1}{|D|}\!\sum_{d'\in D}\!|d^*\!\!\!-\!\!d'|$        & SF: $\!\frac{1}{|SF|}\!\sum_{sf'\in SF}\!|sf^*\!\!\!-\!\!sf'|$         \\ \hline
pixel acc.  $\!\frac{\sum_i n_{ii}}{\sum_i t_i}$ & mean acc.    $\!\frac{1}{n_{cl}}\sum_i\frac{n_{ii}}{t_i}$           \\
mean IoU $\!\frac{1}{n_{cl}}\sum_i\frac{n_{ii}}{t_i+\sum_j n_{ji}+n_{ii}}$ & f.w. IoU: $\!\frac{1}{\sum_i t_i}\sum_i\frac{n_{ii}}{t_i+\sum_j n_{ji}+n_{ii}}$ \\ \hline
\end{tabular}
\label{metrics}
\vspace{-1.8\baselineskip}
\end{table}

\vspace{-0.3\baselineskip}
\subsection{Depth evaluation}
\vspace{-0.3\baselineskip}
\noindent\textbf{Experiment setup.}
The depth experiments are conducted on KITTI 2015 and Cityscapes. For KITTI test split, the given depth ground truth is used for evaluation. For Eigen test split, synchronized Velodyne points are provided and these sparse points are projected and serve as depth ground truth. Only pixels with ground truth depth values are evaluated.
The following evaluations are performed to present the depth results: (1) ablation study of our approach; (2) depth estimation performance comparison with SOTA methods.

\vspace{0.5\baselineskip}
\noindent\textbf{Ablation study.}
We explore the effectivness of each component in our framework. Several variant results are generated for evaluation, which include: (1) DepthNet trained with only monocular training sequences (Ours (mono)); (2) DepthNet trained with monocular samples and then finetuned with HMP (Ours (mono+HMP)); (3) DepthNet without finetuning from 3D solver loss (Ours w/o HMP). For traning with only monocular sequences, the left and right sequences are considered independently, thus resulting in 44,000 training samples. The quantitative results of different variants are presented in Tab. \ref{tbl:sota}. Although these three variants use the same amount of data, our approach trained with both stereo and sequential samples shows large performance boost over using only one type of training samples, proving the effectiveness of incorporating stereo into training. With the finetuning from HMP, the performance is further improved. 

\vspace{0.5\baselineskip}
\noindent\textbf{Comparison with state-of-the-art.}
Following the tradition of other methods \cite{eigen2014depth,zhou2017unsupervised,godard2016unsupervised}, our framework is trained with two strategies: (1) trained with KITTI data only; (2) trained with Cityscapes data and then finetuned with KITTI data (CS+K). The maximum of depth estimation on KITTI split is capped at 80 meters and the same crop as in \cite{eigen2014depth} is applied during evaluation on Eigen split. 

\begin{figure*}
\vspace{-0.5\baselineskip}
\centering
\includegraphics[width=\textwidth]{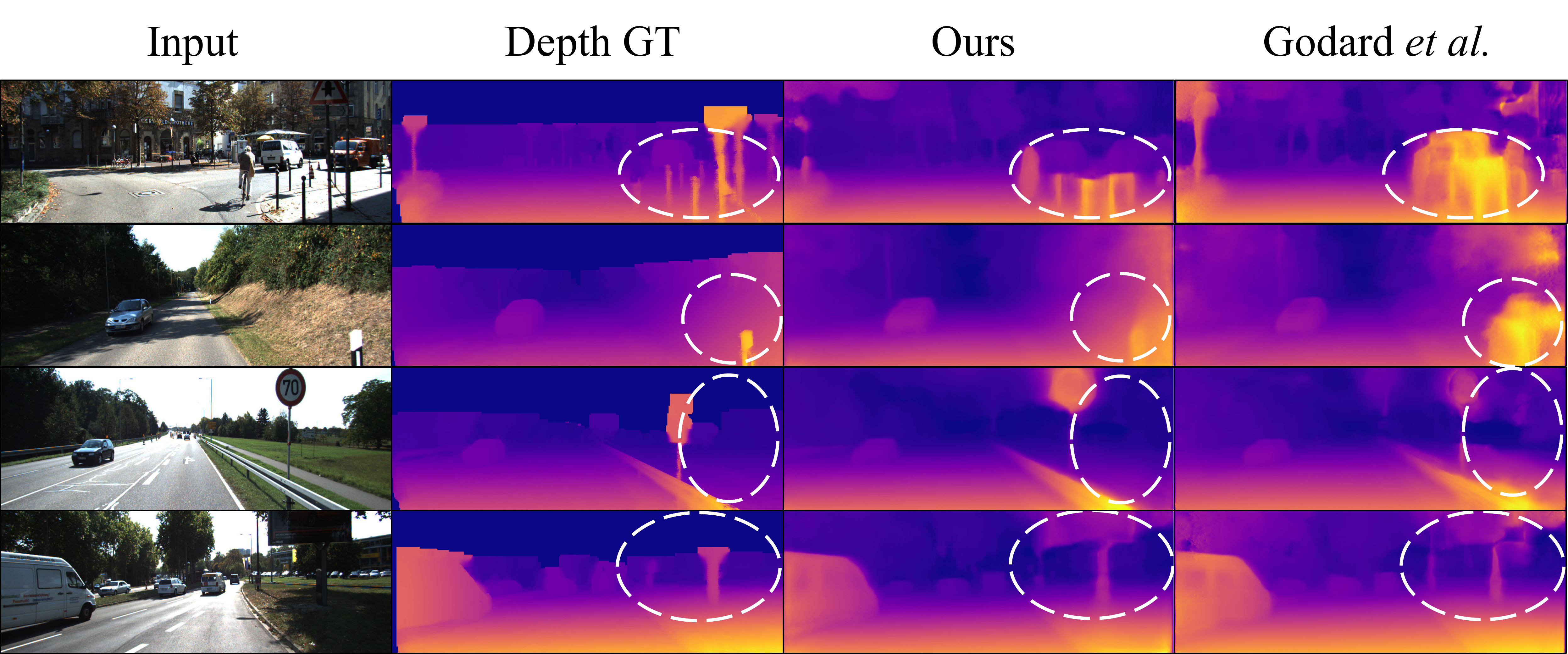}
\caption{Visual comparison between Godard \etal \protect\cite{godard2016unsupervised} and our results on KITTI test split. The depth ground truths are interpolated and all images are reshaped for better visualization. For depths, our results have preserved the details of objects noticeably better (as in white circles).}
\label{fig:examples}
\vspace{-0.8\baselineskip}
\end{figure*}

Tab. \ref{tbl:sota} shows the comparison of ours performance and recent SOTA methods. Our approach outperforms current SOTA unsupervised methods \cite{zhou2017unsupervised,kuznietsov2017semi,yang2018cvpr,godard2016unsupervised} on almost all metrics by a large margin when trained with KITTI data. When trained with more data (CS+K), our method still shows the SOTA performance on the ``Abs Rel'' metric. Some depth estimation visualization results are presented in \figref{fig:example}, comparing with results from \cite{godard2016unsupervised}. Our depth results have preserved the details of the scene noticeably better.

\begin{table}[t]
\vspace{-0.5\baselineskip}
\centering
\caption{Monocular depth evaluation results on KITTI split (upper part) and Eigen split(lower part). Results of \cite{zhou2017unsupervised} on KITTI test split are generated by training their released model on KITTI dataset. All results are generated by model trained on KITTI data only unless specially noted. ``pp'' denotes post processing implemented in \cite{godard2016unsupervised}.}
\label{tbl:sota}
\fontsize{7.5}{9}\selectfont
\bgroup
\def\arraystretch{1.15}
\begin{tabular}{lclcccccccccc}
\specialrule{.2em}{.1em}{.1em}
\multirow{2}{*}{Method}               & \multirow{2}{*}{Split}                 &  & \multirow{2}{*}{Stereo} &  & \multicolumn{4}{c}{Lower the better}                     &  & \multicolumn{3}{c}{Higher the better}                                                                     \\ \cline{6-9} \cline{11-13} 
                                      &                                             &  &                         &  & Abs Rel        & Sq Rel         & RMSE  & RMSE log       &  & $\delta < 1.25$ & $\delta < 1.25^2$ & $\delta < 1.25^3$ \\ \cline{1-9} \cline{11-13} 
\multicolumn{1}{l|}{Train mean}       & \multicolumn{1}{c|}{\multirow{9}{*}{KITTI}} &  &                         &  & 0.398          & 5.519          & 8.632 & 0.405          &  & 0.587                 & 0.764                                   & 0.880                                   \\
\multicolumn{1}{l|}{Zhou \etal\cite{zhou2017unsupervised}}       & \multicolumn{1}{c|}{}                       &  &                         &  & 0.216          & 2.255          & 7.422 & 0.299          &  & 0.686                 & 0.873                                   & 0.951                                   \\
\multicolumn{1}{l|}{LEGO\cite{yang2018cvpr}}             & \multicolumn{1}{c|}{}                       &  &                         &  & 0.154          & 1.272          & 6.012 & 0.230          &  & 0.795                 & 0.932                                   & 0.975                                   \\
\multicolumn{1}{l|}{Wang \etal\cite{wang2017learning}}       & \multicolumn{1}{c|}{}                       &  &                         &  & 0.151          & 1.257          & 5.583 & 0.228          &  & 0.810                 & 0.936                                   & 0.974                                   \\
\multicolumn{1}{l|}{Godard \etal\cite{godard2016unsupervised}}     & \multicolumn{1}{c|}{}                       &  & \checkmark              &  & 0.124          & 1.388          & \textbf{6.125} & 0.217          &  & 0.841                 & 0.936                                   & 0.975                          \\
\multicolumn{1}{l|}{Ours (mono)}             & \multicolumn{1}{c|}{}                       &  &               &  & 0.137 & 1.326 & 6.232 & 0.224 &  & 0.806        & 0.927                          & 0.973 \\
\multicolumn{1}{l|}{Ours (mono+HMP)}             & \multicolumn{1}{c|}{}                       &  &             &  & 0.131 & 1.254 & 6.117 & 0.220 &  & 0.826        & 0.931                         & 0.973 \\
\multicolumn{1}{l|}{Ours (w/o HMP)}             & \multicolumn{1}{c|}{}                       &  & \checkmark              &  & 0.117 & 1.163 & 6.254 & 0.212 &  & 0.849        & 0.932                          & 0.975 \\
\multicolumn{1}{l|}{Ours}             & \multicolumn{1}{c|}{}                       &  & \checkmark              &  & \textbf{0.109} & \textbf{1.004} & 6.232 & \textbf{0.203} &  & \textbf{0.853}        & \textbf{0.937}                          & \textbf{0.975}                          \\ \hline
\multicolumn{1}{l|}{Godard \etal\cite{godard2016unsupervised} (CS+K+pp)}     & \multicolumn{1}{c|}{}                       &  & \checkmark              &  & 0.100          & \textbf{0.934}          & \textbf{5.141} & \textbf{0.178}          &  & \textbf{0.878}                 & \textbf{0.961}                                   & 0.986 \\
\multicolumn{1}{l|}{Ours (CS+K)}             & \multicolumn{1}{c|}{}                       &  & \checkmark              &  & \textbf{0.099} & 0.986 & 6.122 & 0.194 &  & 0.860        & 0.957                          & \textbf{0.986}                          \\ \hline
\multicolumn{1}{l|}{Train mean}       & \multicolumn{1}{c|}{\multirow{7}{*}{Eigen}} &  &                         &  & 0.403          & 5.530          & 8.709 & 0.403          &  & 0.593                 & 0.776                                   & 0.878                                   \\
\multicolumn{1}{l|}{Zhou \etal\cite{zhou2017unsupervised}}       & \multicolumn{1}{c|}{}                       &  &                         &  & 0.208          & 1.768          & 6.856 & 0.283          &  & 0.678                 & 0.885                                   & 0.957                                   \\
\multicolumn{1}{l|}{UnDeepVO\cite{li2017undeepvo}}         & \multicolumn{1}{c|}{}                       &  & \checkmark              &  & 0.183          & 1.730          & 6.570 & 0.268          &  & -                     & -                                       & -                                       \\
\multicolumn{1}{l|}{LEGO\cite{yang2018cvpr}}             & \multicolumn{1}{c|}{}                       &  &                         &  & 0.162          & 1.352          & 6.276 & 0.252          &  & 0.783                 & 0.921                                   & 0.969                                   \\
\multicolumn{1}{l|}{Mahjourian \etal \cite{mahjourian2018unsupervised}} & \multicolumn{1}{c|}{}                       &  &                         &  & 0.163          & 1.240          & 6.220 & 0.250          &  & 0.762                 & 0.916                                   & 0.968                                   \\
\multicolumn{1}{l|}{Godard \etal\cite{godard2016unsupervised}}     & \multicolumn{1}{c|}{}                       &  & \checkmark              &  & 0.148          & 1.344          & \textbf{5.927} & 0.247          &  & 0.803                 & 0.922                                   & 0.964                                   \\
\multicolumn{1}{l|}{Ours}             & \multicolumn{1}{c|}{}                       &  & \checkmark              &  & \textbf{0.127} & \textbf{1.239} & 6.247 & \textbf{0.214} &  & \textbf{0.847}        & \textbf{0.926}                          & \textbf{0.969}                          \\ \hline
\multicolumn{1}{l|}{Godard \etal\cite{godard2016unsupervised} (CS+K+pp)}     & \multicolumn{1}{c|}{}                       &  & \checkmark              &  & 0.118          & \textbf{0.923}          & \textbf{5.015} & 0.210          &  & \textbf{0.854}                 & \textbf{0.947}                                   & 0.976                                   \\
\multicolumn{1}{l|}{Ours (CS+K)}     & \multicolumn{1}{c|}{}                       &  & \checkmark              &  & \textbf{0.114}          & 1.074          & 5.836 & \textbf{0.208}          &  & 0.856                 & 0.939                                   & \textbf{0.976}                                   \\ \hline
\end{tabular}
\egroup
\vspace{-0.5\baselineskip}
\end{table}

\vspace{-0.3\baselineskip}
\subsection{Scene flow evaluation}
\vspace{-0.3\baselineskip}
\label{sf_exp}
\noindent\textbf{Experiment setup.}
The scene flow evaluation is performed on KITTI 2015 dataset. For 200 frames pairs in KITTI test split, the depth ground truth of the two consecutive frames ($t$ and $t+1$) and the 2D optical flow ground truth from frame $t$ to frame $t+1$ are provided. Following the KITTI benchmark evaluation toolkit, the scene flow evaluation is conducted on the two depth results and optical flow results. As the unsupervised method generates depth/disparity up to a scale, we rescale the depth estimation by a factor to make the estimated depth median equal to ground truth depth median.


\vspace{0.5\baselineskip}
\noindent\textbf{Ablation study.}
We explore the effectiveness of HMP and other loss terms by several ablation experiments: (1)excluding the HMP module from our framework (Ours w/o HMP); (2) DepthNet trained with monocular samples (Ours (mono)). The scene flow evaluation results of different variants are presented in Tab. \ref{tbl:sf_sota}. As the same trend in depth evaluation, both incorporating stereo examples into training and finetuning with HMP help improve the scene flow performance.

\begin{figure*}
\vspace{-0.8\baselineskip}
\centering
\includegraphics[width=\textwidth]{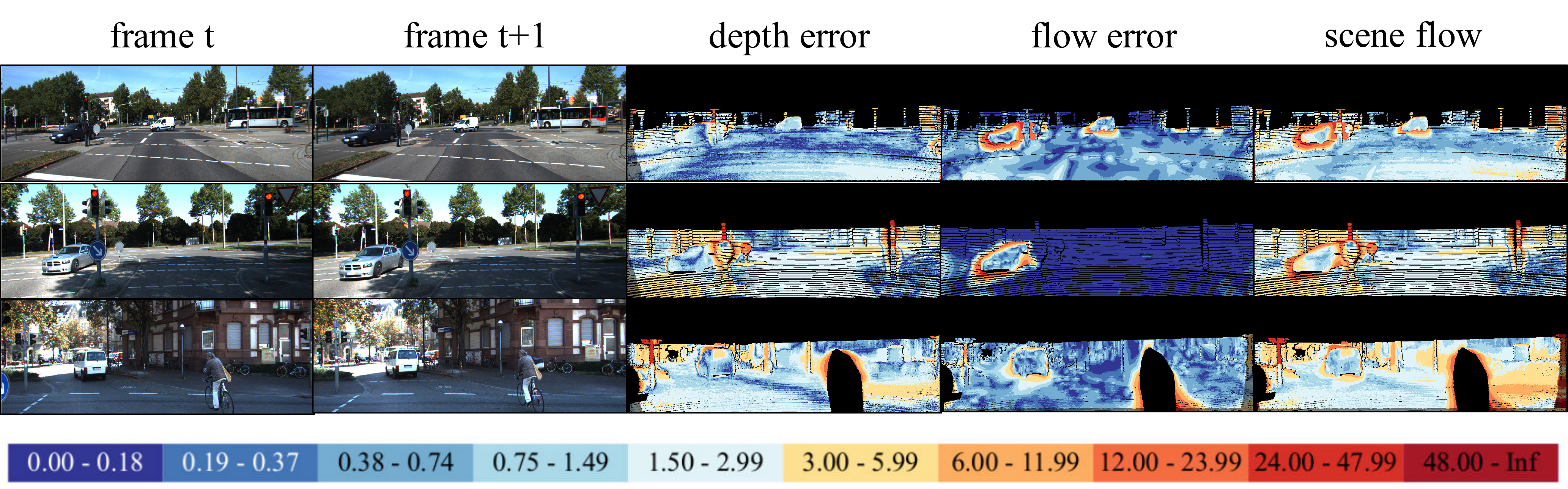}
\caption{Errors in scene flow evaluation. The left two columns show the two consecutive frames as input. The other three columns show the error in depth, flow and scene flow evaluation. The color code of error is following the tradition of \cite{menze2015cvpr}.}
\label{fig:sf_error}
\vspace{-1.0\baselineskip}
\end{figure*}

\vspace{0.5\baselineskip}
\noindent\textbf{Comparison with other methods.}
The comparison with current SOTA scene flow methods are presented in Tab. \ref{tbl:sf_sota}. Note that all supervised methods use the stereo image pairs to generate the disparity estimation during testing. 
The performance of ``Ours w/o HMP'' is further improved with scene flow solver, proving the capability of facilitating depth learning through optical flow in the proposed HMP. The depth, flow and scene flow errors are visualized in Fig. \ref{fig:sf_error}.

\begin{table}[t]
\vspace{-0.5\baselineskip}
\centering
\caption{Scene flow performances of different methods on KITTI 2015 dataset. Upper part includes results of supervised methods and the bottom part includes unsupervised methods.}
\label{tbl:sf_sota}
\begin{tabular}{lc|ccc|ccc|ccc}
\specialrule{.2em}{.1em}{.1em}

\multicolumn{1}{l|}{}                     & \multirow{2}{*}{Supervision} & \multicolumn{3}{c|}{D1}                    & \multicolumn{3}{c|}{D2}                    & \multicolumn{3}{c}{FL}                     \\
\multicolumn{1}{l|}{}                     &             & \textit{bg} & \textit{fg} & \textit{bg+fg} & \textit{bg} & \textit{fg} & \textit{bg+fg} & \textit{bg} & \textit{fg} & \textit{bg+fg} \\ \hline
\multicolumn{1}{l|}{OSF\cite{menze2015cvpr}}                 & Yes         & 4.00        & 8.86        & 4.74           & 5.16        & 17.11       & 6.99           & 6.38        & 20.56       & 8.55           \\
\multicolumn{1}{l|}{ISF\cite{behl2017bounding}}                 & Yes         & 3.55        & 3.94        & 3.61           & 4.86        & 4.72        & 4.84           & 6.36        & 7.31        & 6.50           \\ \hline
\multicolumn{1}{l|}{Ours w/o HMP}     & No          & 24.22        & 27.74        &26.38           & 68.84        & 71.36       & 69.68           & 25.34        & 28.00       & 25.74          \\
\multicolumn{1}{l|}{Ours(mono)}           & No          & 26.12        & 30.27        & 30.54           & 23.94        & 68.47       & 73.85           & 25.34        & 28.00       & 25.74          \\
\multicolumn{1}{l|}{Ours}         & No          & 23.62        & 27.38        & 26.81           & 18.75       & 60.97       & 70.89           & 25.34        & 28.00       & 25.74           \\ \hline
\end{tabular}
\vspace{-0.5\baselineskip}
\end{table}

\vspace{-1.0\baselineskip}
\subsection{Moving object segmentation}
\vspace{-0.5\baselineskip}
\label{seg_exp}
We evaluate the moving object segmentation performance to test the capability of capturing foreground motion in our framework.

\begin{figure*}
\vspace{-0.8\baselineskip}
\centering
\includegraphics[width=\textwidth]{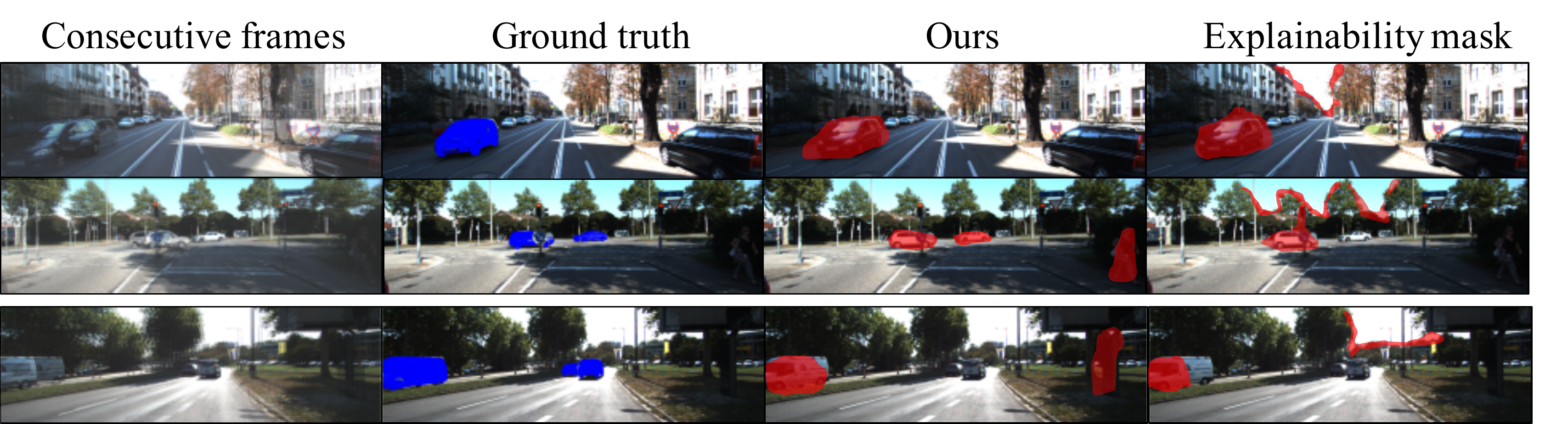}
\caption{Moving object segmentation results.}
\label{fig:seg}
\vspace{-1.8\baselineskip}
\end{figure*}

\vspace{0.5\baselineskip}
\noindent\textbf{Experiment setup.} The moving object segmentation is evaluated on KITTI 2015 dataset. ``Object map'' ground truth is provided in this dataset to dinstinguish foreground and background in flow evaluation. Such dense motion mask serve as ground truth in our segmentation evaluation. Fig. \ref{fig:seg} (second column) shows some visualization of segmentation ground truths. 

For better quantitative comparison, we propose several baseline methods to do moving object segmentation, including: (1) Using segment mask from MotionNet in the same way as explainability mask of \cite{zhou2017unsupervised} with our learning pipeline by removing HMP; 
(2) Compute a residual flow map by substracting 3D flow induced by camera motion (using $\ve{T}_{t\rightarrow s},\ve{D}_t,\ve{V}_t$) from the full 3D scene flow (using $\ve{F}_{t\rightarrow s}, \ve{D}_t, \ve{D}_s, \ve{V}_t$). Then, we apply a two-class Gaussian Mixture Model (GMM) to fit the flow magnitude, on which do graph cut to generate the segmentation results (Graphcut on residual flow). We leave the segmentation details in supplimentary material due to space limit.

\vspace{0.5\baselineskip}
\noindent\textbf{Evaluation results.}
We compare our segmentation results from the motion mask and those from the two baseline methods. As the Tab. \ref{tbl:seg} shows, our segmentation results from the motion mask shows superior performance compared to the masks applied in depth reconstruction or masks calculated from the scene flow residual. Visualization examples of segmentation are presented in Fig. \ref{fig:seg}. Our segmentation results are focused on moving object compared to the explainability masks similar to \cite{zhou2017unsupervised}, which is optimized to filter out any reconstruction error.

\begin{table}[t]
\vspace{-0.5\baselineskip}
\centering
\caption{Foreground moving object segmentation performance on KITTI 2015 dataset.}
\label{tbl:seg}
\begin{tabular}{l|cccc}
\specialrule{.15em}{.08em}{.08em}
                             & pixel acc. & mean acc. & mean IoU & f.w. IoU \\ \hline
Explainability mask & 70.32      & 58.24     & 41.95    & 67.56    \\
Graphcut on residual flow    & 75.05      & 67.26     & 50.83    & 71.32    \\
Ours                         & \textbf{88.71}      & \textbf{74.59}     & \textbf{52.25}    & \textbf{86.53}   \\ \hline
\end{tabular}
\vspace{-0.5\baselineskip}
\end{table}

\vspace{-0.5\baselineskip}
\section{Conclusion} 
\vspace{-0.5\baselineskip}
In this paper, we proposed a self-supervised framework for joint 3D geometry and dense object motion learning. A novel depth estimation framework is proposed to model better depth estimation and also the ego-motion. A holistic 3D motion parser (HMP) is proposed to model the consistency between depth and 2D optical flow estimation. Such consistency is proved to be helpful for supervising depth learning. 
We conducted comprehensive experiments to present the performance. On KITTI dataset, our approach achieves SOTA performance on all depth, scene flow and moving object segmentation evaluations.  In the future, we would like to extend our framework to other motion video data sets containing deformable and articulated non-rigid objects such as MoSeg~\cite{brox2010object} \etc, in order to make the learning as general as possible.


\bibliographystyle{splncs}
\bibliography{unsp_motion_depth}
\end{document}